\documentclass[10pt,twocolumn,letterpaper]{article}

\usepackage{graphicx}
\graphicspath{ {image/} }
\usepackage{amsmath}
\usepackage{color}
\usepackage{multirow}
\usepackage{bm}
\usepackage{epsfig}
\usepackage{graphicx}
\usepackage[switch]{lineno}
\usepackage{amssymb}

\usepackage[breaklinks=true,bookmarks=false]{hyperref}

\def\cvprPaperID{****} 

\begin{document}

\title{Learning Deep Representations for Semantic Image Parsing: \\ a Comprehensive Overview}

\date{} 
\author{Lili Huang, Jiefeng Peng, Ruimao Zhang, Guanbin Li, Liang Lin\\
\\ School of Data and Computer Science, Sun Yat-Sen University}

\maketitle

\begin{abstract}
Semantic image parsing, which refers to the process of decomposing images into semantic regions and constructing the structure representation of the input, has recently aroused widespread interest in the field of computer vision. The recent application of deep representation learning has driven this field into a new stage of development. In this paper, we summarize three aspects of the progress of research on semantic image parsing, i.e., category-level semantic segmentation, instance-level semantic segmentation, and beyond segmentation. Specifically, we first review the general frameworks for each task and introduce the relevant variants. The advantages and limitations of each method are also discussed. Moreover, we present a comprehensive comparison of different benchmark datasets and evaluation metrics. Finally, we explore the future trends and challenges of semantic image parsing.

\end{abstract}

\section{Introduction}
\subsection{Semantic Image Parsing}

\noindent With the development of Internet, in recent years, large-scale image and multimedia video data have increased explosively, resulting in urgent demands for advanced intelligent image analysis technology, such as semantic image parsing. As a fundamental and long-standing problem in computer vision, semantic image parsing is performed at three levels, which will be discussed below.

\begin{figure}[t]
\centering
\includegraphics[width= 0.9\linewidth]{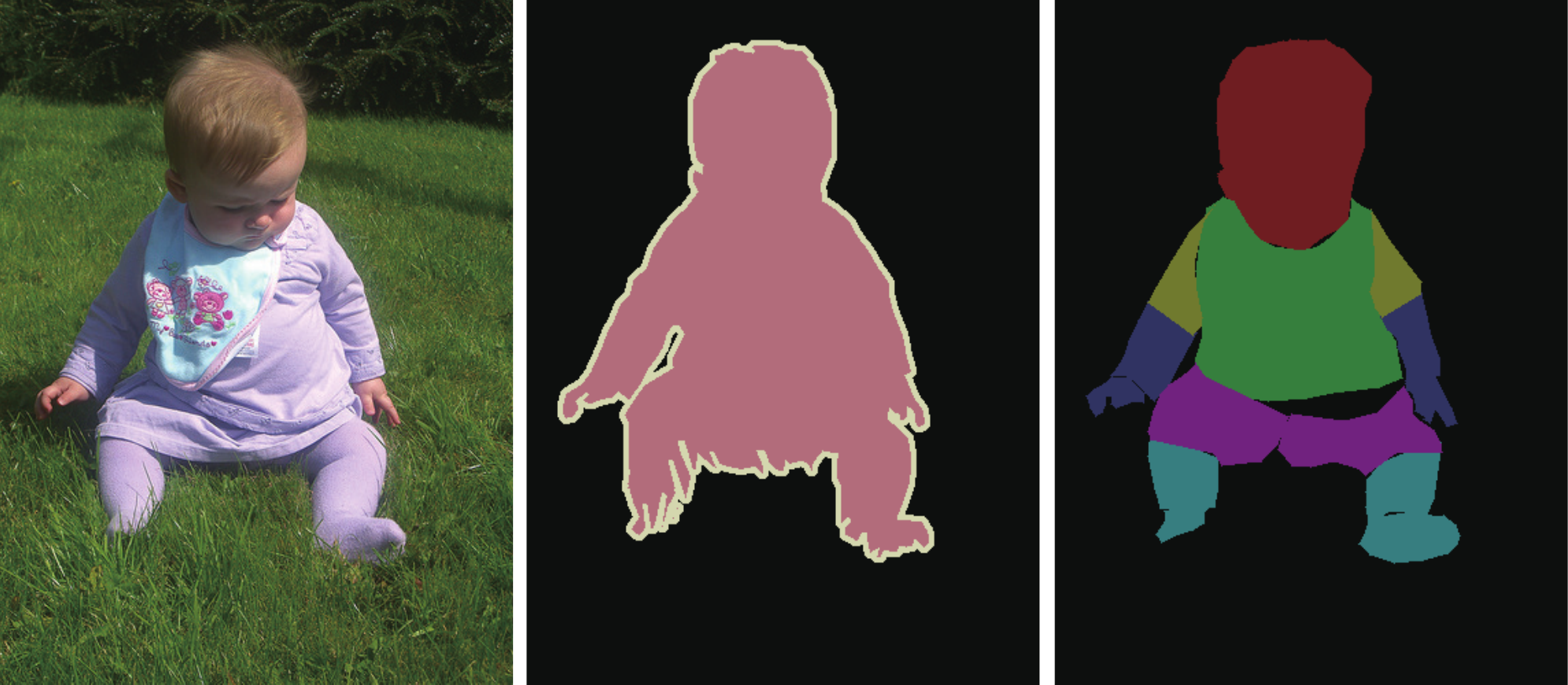}
\caption{Illustration of the category-level semantic segmentation. The left is the original image. The middle is the basic semantic segmentation result, and the right is the semantic part segmentation result.}
\label{fig:1}
\end{figure}

\romannumeral1. \textbf{Category-level semantic segmentation}. It attempts to assign a single category label to each pixel. Here, a category label corresponds to a specific object category or a local part of the object. Therefore, category-level semantic segmentation consists of basic semantic segmentation and semantic part segmentation (called object parsing in the literature), as illustrated in Fig. \ref{fig:1}. The former predicts the segmentation mask and its label for the entire object, as shown in the middle of Fig. \ref{fig:1}, while the latter refers to segmenting an object into its constituent semantic parts and predicting the segmentation mask for each local part, as shown on the right side of Fig. \ref{fig:1}. According to the definition, part segmentation can be regarded as a special type of fine-grained category-level semantic segmentation task.

Category-level semantic segmentation is actually a pixel-wise dense prediction problem, which is supported by two key technologies: 1) classification: an object is assigned a specific semantic-concept label; and 2) localization: the classification label for a pixel must match the appropriate coordinates in the output score map \cite{zhao2016pyramid}.

\begin{figure}[t]
\centering
\includegraphics[width= \linewidth]{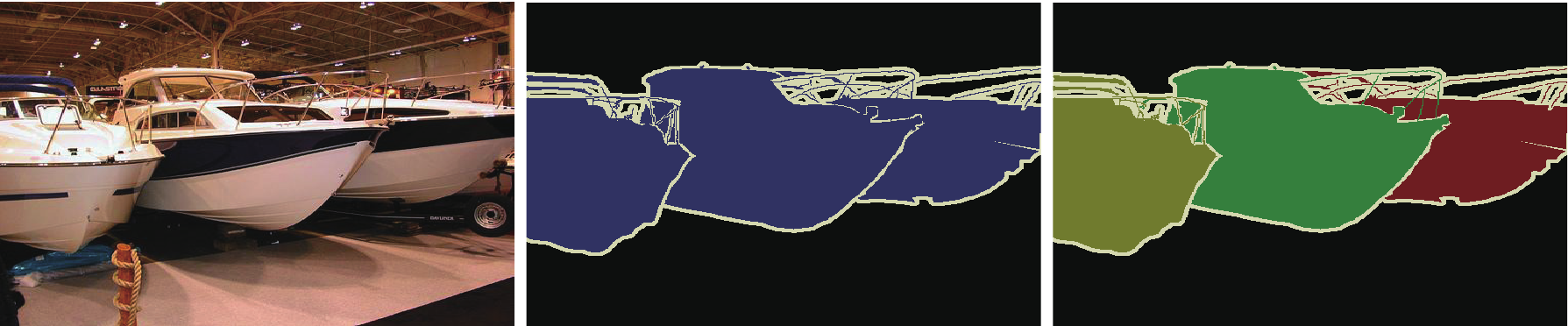}
\caption{Comparison of category-level and instance-level semantic segmentation.The left is the original image. The middle is the category-level semantic segmentation result, and the right is the instance-level semantic segmentation result.}
\label{fig:2}
\end{figure}

\romannumeral2. \textbf{Instance-level semantic segmentation}. In contrast to category-level segmentation, it requires precise segmentation of each object and correct detection of all the object instances in one image \cite{he2017mask}. In the middle of Fig. \ref{fig:2}, three boats are segmented by assigning the same category label (i.e., "boat"). Clearly, category-level segmentation cannot distinguish the object instances belonging to the same category. In the right column of Fig. \ref{fig:2}, the three boats are segmented by assigning different IDs with the same category label (i.e., "boat one", "boat two", and "boat three"). Thus, the instance-level segmentation requires support from both classification and detection technologies.

\begin{figure}[t]
\centering
\includegraphics[width= \linewidth]{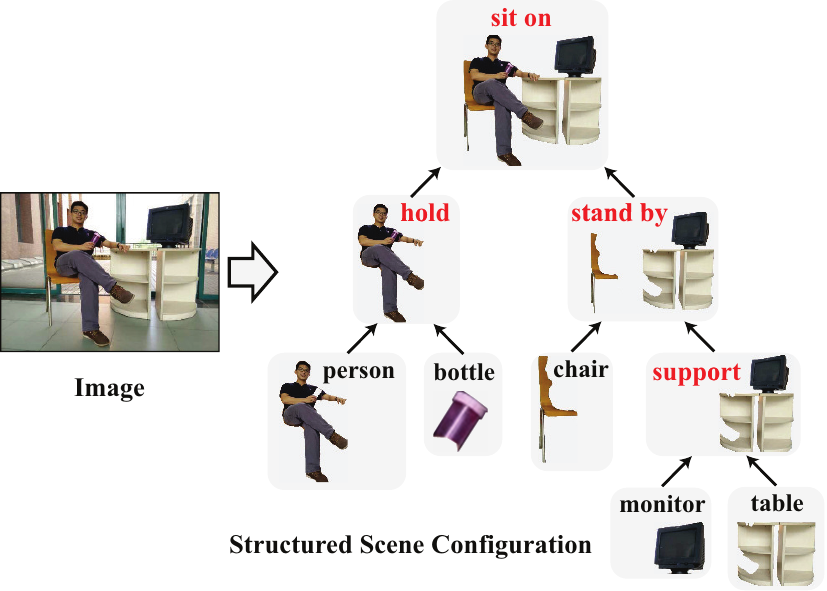}
\caption{Illustration of beyond segmentation. (Figure extracted from \cite{lin2016deep})}
\label{fig:3}
\end{figure}

\romannumeral3. \textbf{Beyond segmentation}. In recent years, works extending beyond semantic segmentation have also received substantial attention. This task is inspired by previous work on image parsing \cite{he2017mask}, which refers to the process of decomposing an image into its constituent visual structured configuration \cite{tu2003image, tu2002parsing, han2005bottom}. Works beyond segmentation not only semantically segment images but also predict richer and finer results, such as the structures and relations of objects and the spatial layout. Specifically, images are decomposed into semantic regions and the structures and relationships among objects are constructed. For example, in Fig. \ref{fig:3}, the image caption is "there is one person sitting on the chair nearby the table with one monitor". Following the work in \cite{lin2016deep}, the beyond segmentation method first segments all the objects (i.e., "person", "chair", "table", and "bottle") in the image, predicts the relations among objects (i.e., "hold", "stand by", "support", and "sit on"), and finally estimates the hierarchical structures. Intuitively, works beyond segmentation produces detailed parsing results that are consistent with human perception.

Similar to most vision problems, the discriminant features greatly affect the performance of semantic image parsing. Traditional semantic segmentation methods adopt hand-crafted features, such as SIFT \cite{lowe2004distinctive}, HOG \cite{dalal2005histograms}, and LBP \cite{ahonen2004face}. However, these hand-crafted features are not applicable to various tasks. Therefore, the automatic extraction of valuable information and effective representation of image/video data are critical. Representation learning, i.e., learning representations of data, makes it easier to extract useful information from raw data to build predictors. The representation algorithms for semantic image parsing have experienced three periods of progress in the continuous improvement of image parsing performance: 1) traditional hand-crafted methods; 2) deep learning, such as convolutional neural networks (CNNs), recurrent neural networks, and recursive neural networks (RNNs); and 3) the integration of the two methods to complement each other.

Extensive experiments \cite{liu2015semantic,chen2016deeplab,long2015fully,chen2014semantic,he2017mask,peng2017large,zhao2016pyramid} have demonstrated that the representation ability of traditional hand-crafted features is insufficient. Meanwhile, deep learning currently achieves the best representation ability and has had tremendous success in many applications, such as image classification \cite{krizhevsky2012imagenet}, object detection, and natural language understanding \cite{socher2010learning}. Therefore, we list only the main differences among the three-level semantic segmentation tasks accomplished by deep representation models, as illustrated in Table \ref{tab:1}.

\doublerulesep 0.1pt
\begin{table*}[t]
\renewcommand\arraystretch{1.5}
\begin{footnotesize}

\begin{center}
\resizebox{\textwidth}{11mm}{
\begin{tabular}{c|cccc}
\hline\hline\noalign{\smallskip}
    Task    &Flourshing period        &  Pioneering work          &Key technology & Type of labels   \\
\noalign{\smallskip} \hline
   Category-level segmentation              & 2015 & FCN \cite{long2015fully}   & classification, localization & object, part \\
   Instance-level segmentation        & 2016    & FCIS \cite{li2016fully} & classification, detection & instance \\
   Beyond segmentation     & 2016      & CNN-RNN \cite{lin2016deep}  & classification, localization & object, part, relation, scene structure\\
\hline\hline
\end{tabular}}
\caption{Comparisons of different semantic segmentation tasks performed by deep models} \label{tab:1}
\end{center}
\end{footnotesize}
\end{table*}

\subsection{Deep Learning}

Deep learning is defined as learning multiple levels of representations from the local and detailed levels in the shallow layers to the global and abstract levels in the deeper layers \cite{Bengio2013Deep, bengio2012deep, bengio2013representation}. Specifically, deep neural networks consist of several simple but non-linear modules, each of which transforms the simple representation at the shallow layer (starting with the raw input) into slightly more abstract representation at the deep layer. Several well-known deep neural networks, such as the CNN, recurrent neural network, and RNN, have been reported in recent years. Moreover, abundant variants of these networks, which we discuss in the following sections, have emerged.

\textbf{Convolutional Neural Networks}. The CNN \cite{lecun1989backpropagation} is designed for data with grid-like structures and consists of convolutional layers, pooling layers, and non-linear rectification layers. The units in the neural network are locally connected, which results in shared weights of the local parameters and features in the deeper abstract layers being invariant to local image transformation. Despite the numerous applications of CNNs, they were not well-known until their successful application to object recognition during the ImageNet challenge in 2012. Then, CNN was quickly applied to semantic segmentation \cite{long2015fully, dai2016instance, he2017mask, peng2017large, zhao2016pyramid, li2016fully, islam2017label} and achieved great successes.

\textbf{Recurrent Neural Networks}. In contrast to CNNs, which are tailored for grid-structure data \cite{lecun1989backpropagation}, recurrent neural networks are more appropriate for sequential data \cite{Lipton2015A}. The principal characteristic of a recurrent neural network is that neurons (units) are connected by synaptic links to express temporal relations. To alleviate the explosion or vanishing of the backpropagated gradients in the shallow layers \cite{lecun2015deep, Lipton2015A}, long short-term memory (LSTM) networks \cite{liang2016semanticA} were proposed by introducing special hidden units to memorize the observed knowledge of the previous and current inputs. The success of LSTM has demonstrated that LSTM is more effective than conventional recurrent neural networks in image captioning \cite{karpathy2015deep} and machine translation \cite{sutskever2014sequence}. Additionally, many works \cite{li2016lstm, liang2016semanticA, peng2016geometric, byeon2015scene, liang2016semanticB, liang2017interpretable, zhang2017progressively} utilize LSTM to improve the performance of semantic image parsing.

\textbf{Recursive Neural Networks}. Unlike the aforementioned recurrent neural networks \cite{elman1991distributed}, which are designed for time sequential data, RNNs \cite{socher2010learning} are designed for hierarchical space structural data. Recurrent neural networks for chain structures by connecting hidden units, whereas RNNs recursively form a hierarchical structure because the structures of networks are similar at every level of the hierarchy. This characteristic is in line with the structures of natural language, which results in successful natural language parsing \cite{socher2010learning}. Some recent works \cite{lin2016deep,socher2010learning} proposed RNNs for structural semantic parsing.

\subsection{Our Contribution to the Existing Surveys}

With a unique perspective, this work comprehensively reviews deep representation learning-based semantic image parsing at three levels: category-level semantic segmentation, instance-level semantic segmentation, and beyond segmentation. Specifically, for each level of semantic segmentation, we elaborate the relative terminology and background knowledge. Furthermore, this paper reviews and compares existing models and relatively well-known datasets and evaluation metrics. To the best of our knowledge, there is no such overview of semantic image parsing in the literature.

The rest of this article is organized as follows. In Section 2, we review deep representations for semantic image parsing at three levels. Datasets and evaluation metrics are introduced in Section 3. Finally, we present the conclusions and discuss promising future research directions in Section 4.

\section{Learning Deep Representations}
In previous decades, most of the successful semantic segmentation algorithms have relied on hand-crafted features combined with flat classifiers, such as boosting \cite{liu2015parsenet} and support vector machines \cite{szegedy2015going}. Nevertheless, the performance of these algorithms is compromised by the limited feature expression.

More recently, with the emergence of big data and development of computer hardware, deep neural networks have reached their prime. In the field of computer vision, deep leaning has achieved great success in image classification \cite{long2015fully,he2016deep,krizhevsky2012imagenet,lecun1989backpropagation,simonyan2014very}, recurrent neural networks have made tremendous achievements in expressing temporal relations \cite{pinheiro2014recurrent,Graves2007Multi,byeon2015scene, elman1991distributed, Lin2017Knowledge}, and RNNs have succeeded in terms of space structure relationship representation \cite{socher2010learning, lin2016deep}. The breakthroughs of deep learning in image classification are quickly repurposed to semantic image parsing. We illustrate this problem at different levels of image segmentation, i.e., category-level semantic segmentation, instance-level semantic segmentation, and beyond segmentation, in the following sections.

\subsection{Category-Level Semantic Segmentation}

As mentioned in Section 1, category-level semantic segmentation attempts to assign a single category label to each pixel, i.e., basic semantic segmentation and semantic part segmentation, as illustrated in Fig. \ref{fig:1}. For convenience, we do not distinguish between these two processes. The category-level deep models for semantic segmentation are mainly divided into two types: region-based networks and fully convolutional frameworks. 

\textbf{Region-based Networks}. The previously reported deep models \cite{farabet2013learning, gupta2014learning, ning2005toward} are mainly region-based networks that classify each pixel by using its enclosing region for training and prediction. These methods have several limitations. First, they treat each region or pixel as a separate unit. On the one hand, this treatment ignores the importance of the context information in pixel labeling inference, while on the other hand, it ignores the spatial correlation in the image and reduces the algorithm accuracy. Secondly, the independent processing of thousands of regions results in substantial overhead and inefficiency. 

\textbf{Fully Convolutional Frameworks}. The fully convolutional frameworks for semantic segmentation consist of two fundamental works, i.e., fully convolutional networks (FCNs) \cite{long2015fully} and the DeepLab system \cite{chen2014semantic}, both of which fully utilize convolutional networks to produce spatially dense predictions.

\begin{figure}[t]
\centering
\includegraphics[width= 0.9\linewidth]{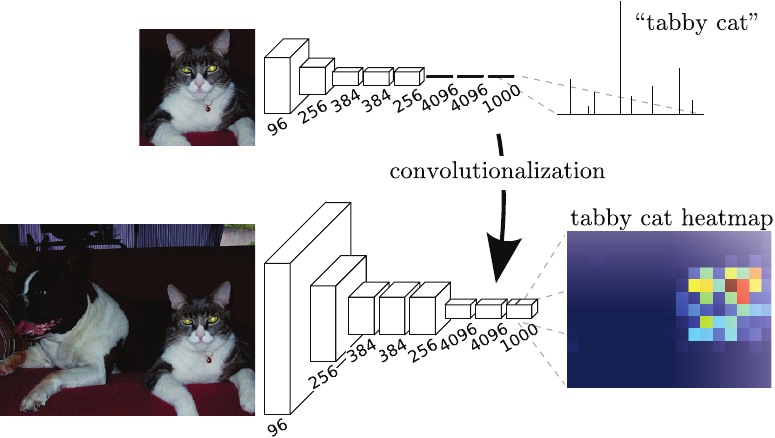}
\caption{Illustration of the adaptation of fully connected layers into convolutional layers. (Figure extracted from \cite{long2015fully})}
\label{fig:5}
\end{figure}

\begin{figure*}[t]
\centering
\includegraphics[width= 0.9\linewidth]{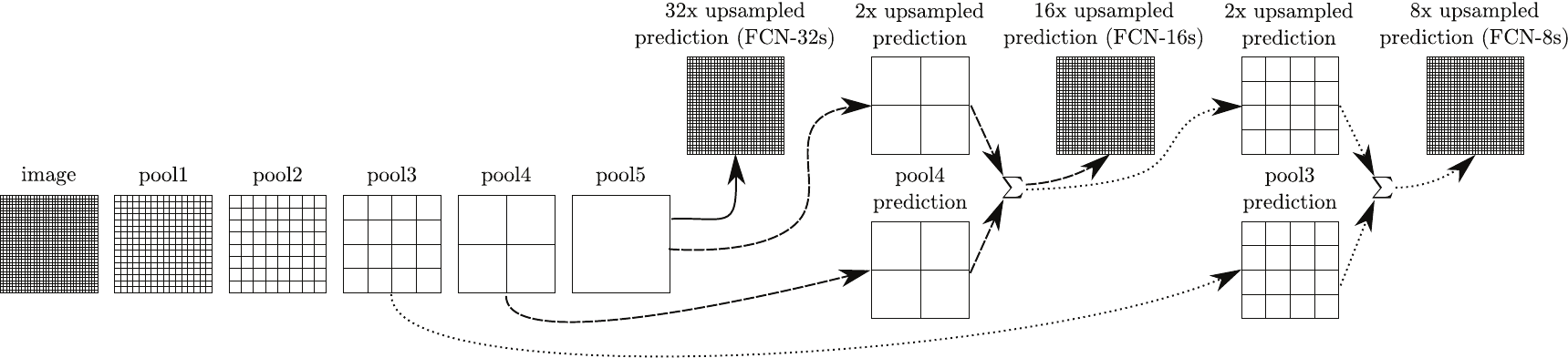}
\caption{The "skip" architecture of FCN. (Figure extracted from \cite{long2015fully})}
\label{fig:6}
\end{figure*}

In \cite{long2015fully}, as the fundamental application of the CNN architecture \cite{liang2015deep, liang2015human}, the authors devised FCNs for spatially dense prediction tasks by accommodating the prior advance deep networks \cite{krizhevsky2012imagenet, simonyan2014very, szegedy2015going}. Specifically, as illustrated in Fig. \ref{fig:5}, fully connected layers in prior networks are converted into convolutional layers, and the deconvolutional layers are built by upsampling intermediate feature maps to keep the size of the output the same as that of the input image. However, the spatial resolution of the feature maps is reduced after the consecutive combination of the max-pooling and downsampling layers in FCN, as in prior image classification models. A novel skip architecture was devised to fuse semantic information with appearance information to produce accurate and detailed segmentations \cite{long2015fully}, as shown in Fig. \ref{fig:6}. The semantic information comes from a deep, coarse layer, while the appearance information is from a shallow, fine layer. Thus, by utilizing prior image classification models as pre-trained models, FCN is fine-tuned to learn and inference efficiently in an end-to-end manner, resulting in equivalently sized output.

\begin{figure}[t]
\centering
\includegraphics[width= 0.9\linewidth]{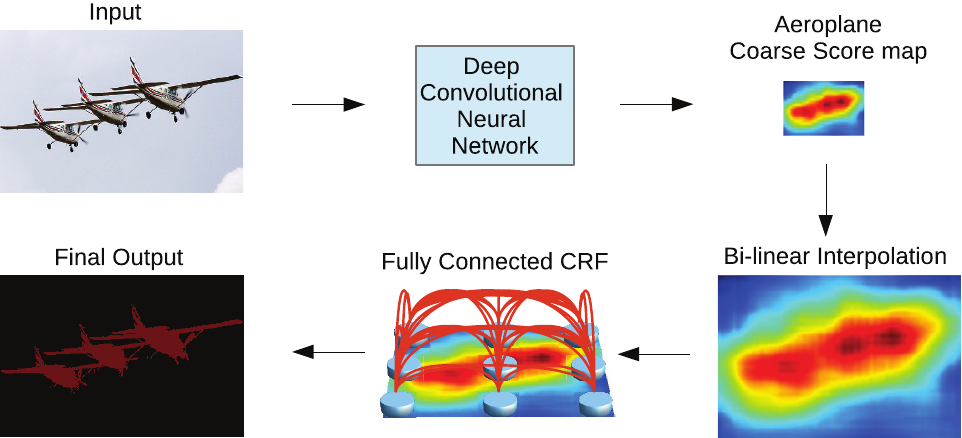}
\caption{Illustration of the DeepLab system. (Figure extracted from \cite{chen2014semantic})}
\label{fig:7}
\end{figure}

Another fundamental work$--$DeepLab system \cite{chen2014semantic}$--$
integrated CNN with fully connected conditional random field (CRF) to expand and improve FCN \cite{long2015fully}. As shown in Fig. \ref{fig:7}, the responses at the final convolutional layer are fed into the fully connected CRF to capture finer details. Thus, the fully connected CRF refines the raw CNN scores, especially along object boundaries. However, the DeepLab system treats CNN and CRF as two separate components. Concretely, fully connected CRF utilizes the Gaussian CRF potentials \cite{koltun2011efficient} to capture long-range dependencies by treating every pixel as a CRF node to receive unary potentials.

Many subsequent variants emerges from these two fundamental works. These works generally evolve along three directions: CNN crafting tricks, integration with the random field model, and integration with recurrent neural networks. We discuss these three aspects below.

\subsubsection{CNN Crafting Tricks}

The majority of deep learning algorithms are based on CNNs; therefore, one intuitive fundamental idea is to design more efficient network architecture with CNN crafting tricks, such as downsample-upsample operation, pyramid module, skip connection, and atrous convolution.

\textbf{Downsample-Upsample Operations.} A downsample-upsample operation is composed of two stages: downsampling and upsampling. In the downsampling stage, the feature maps are processed by convolution or unpooling and progressively shrink to smaller maps, where the receptive field of every pixel is gradually enlarged. In the upsampling stage, the object spatial dimension is recovered through deconvolution or unpooling, where the coarse-to-fine details are captured.

DeconvNet \cite{noh2015learning} treats the convolutional layers of the VGG 16-layer net as the downsampling stage, whereas the developed deconvolution network serves as the upsampling stage, which consists of deconvolution and unpooling layers to increase the resolution of small score maps with more detailed structures. Specifically, DeconvNet first generates sufficient instance-wise candidate proposals for each given image at the downsampling stage, and produces the semantic segmentation maps of each proposal at the upsampling stage. Then, the final semantic segmentation of the whole input image is obtained by assembling the maps of all proposals with non-maximum suppression. Furthermore, DeconvNet \cite{noh2015learning} is integrated with FCN \cite{long2015fully} to improve the performance.

Similar to DeconvNet \cite{noh2015learning}, SegNet \cite{badrinarayanan2015segnet} also introduces the unpooling operation without ReLU in the upsampling stage to recover the spatial dimensions, and the downsampling and upsampling correspond to encoder and decoder stacks, respectively. Specifically, the encoder stacks, composed of convolutions, ReLU and max-pooling, produce low-resolution feature maps while simultaneously memorizing the pooled indices. Then, the decoder stacks upsample the low-resolution maps using the pooled indices and output the semantic segmentation.

\begin{figure*}[t]
\centering
\includegraphics[width= 0.9\linewidth]{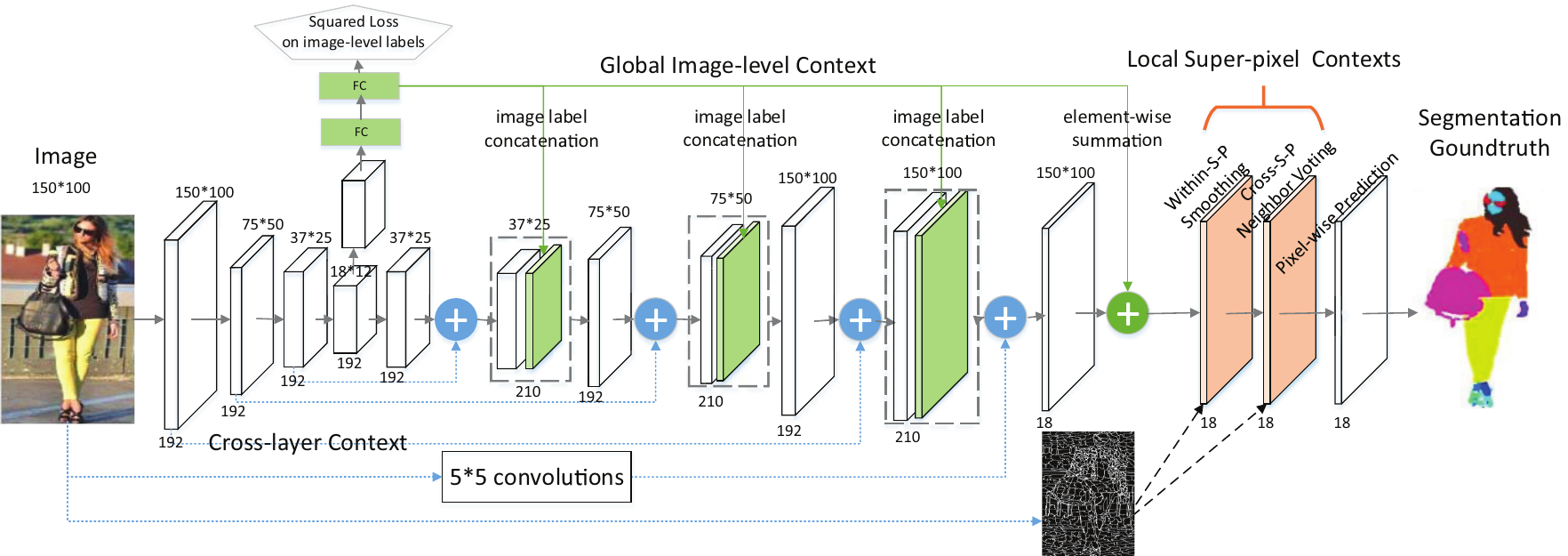}
\caption{Illustration of the Co-CNN. (Figure extracted from \cite{liang2015human})}
\label{fig:8}
\end{figure*}

The more complicated contextualized convolutional neural network (Co-CNN) \cite{liang2015human} is a novel downsample-upsample framework that simultaneously captures hierarchical information by seamlessly integrating three levels of context (i.e., cross-layer context, global image-level context, local super-pixel context) into a unified network, as shown in Fig. \ref{fig:8}. Specifically, Co-CNN first utilizes convolutional networks to obtain the downsampled feature maps for multiple resolutions, upsamples the feature maps along with multi-level context generation, and finally produces pixel-wise predictions. Moreover, the cross-layer context, global image-level context and local super-pixel context are generated by integrating the hierarchical structure, predicting the global image-level labels, and refining super-pixels, respectively.

In general, downsampling is used to extract features from the input image, whereas upsampling produces object segmentation from the features extracted by downsampling. The seamlessly integration of downsampling with upsampling elegantly accomplishes the semantic segmentation task.

\textbf{Pyramid Module.} The pyramid module consists of two varieties: 1) input pyramid, where multi-scale inputs are fed into the same model with shared weights such that the large-scale inputs maintain more fine details and the small-scale inputs capture longer range information; and 2) pooling pyramid, where context information is captured by spatial pyramid pooling in several ranges.

DeepLabV2 \cite{chen2016deeplab}, the updated DeepLab system \cite{chen2014semantic}, employs both types of pyramid modules. On the one hand, DeepLabV2 first transforms the inputs into several scale inputs that are synchronously fed into the weight-shared CNN to produce multi-scale feature maps, which are then merged. On the other hand, DeepLabV2 segments objects at multiple scales via atrous spatial pyramid pooling (ASPP), which serves to resample features prior to convolution. Specifically, ASPP takes advantage of several parallel atrous convolutions with diverse sampling rates to capture multi-scale objects and image context.

Zhao et al. proposed a superior framework$--$pyramid scene parsing network (PSPNet) \cite{zhao2016pyramid}$--$for scene parsing in complex scenes. PSPNet \cite{zhao2016pyramid} adopts the pyramid pooling module to capture the global context representation and prevent the loss of context information between subregions. Specifically, the pyramid pooling module employs multiple pyramid scales to generate coarse to fine feature maps, which provide additional multi-scale contextual information from different regions. Then, the different-region-based context information is aggregated to capture the global context representation.

Apparently, the pyramid module captures multi-scale context information from local fine to global abstract to improve the performance of semantic segmentation.

\textbf{Skip Connection.} Similar to its first application in FCN \cite{long2015fully}, skip connection refers to the links between low-level layers and high-level layers at an interval of several layers; thus, detailed appearance features from shallow layers are combined with coarse semantic information from deep layers to improve the segmentation performance.

The global convolutional network (GCN) \cite{peng2017large} with large-size kernels utilizes pretrained ResNet \cite{he2016deep} as the feature network and FCN \cite{long2015fully} as the segmentation framework. Specifically, the GCN and boundary refinement block are both treated as residual structures. In the feature network, each stage of the ResNet block generates different-scale feature maps, which are fed into the GCN structures to produce semantic score maps for each category. Additionally, the boundary refinement blocks are used to further refine the object boundaries. Next, outputs from the top layer of the residual structures are passed to the segmentation framework, and new high-resolution score maps are generated iteratively by skip connection \cite{long2015fully}. Specifically, upsampled score maps in the higher layers are iteratively combined with the corresponding-resolution score maps extracted from the residual structures in the lower layers. Finally, the semantic score map, which is used to output pixel-wise semantic labels, is generated after the last upsampling.

On the basis of FCN \cite{long2015fully}, U-Net \cite{ronneberger2015u} proposed a u-shaped architecture composed of a contracting path and a symmetric expanding path to effectively train deep models on small datasets. Specifically, the contracting path is similar to the typical convolution architecture used to extract and downsample the feature maps. The lowest-resolution feature maps flows into the expanding path, where the feature maps at each step are upsampled and concatenated with the same resolution feature maps cropped from the contracting path. Thus, the final segmentation maps for each category are generated after the top layer in the expanding path. The cropping step is applied to prevent the loss of border pixels during convolution operations.

Islam et al. \cite{islam2017label} proposed the label refinement network (LRN) to improve segmentation performance by predicting segmentation labels at multiple resolutions. The LRN is formulated as an encoder-decoder framework \cite{long2015fully,badrinarayanan2015segnet,noh2015learning}, where the VGG16 network serves as the encoder network to extract feature maps with decreasing resolution and the decoder network predicts multi-scale coarse-to-fine label maps in several stages. The skip connection architecture combines the label maps of each stage with the corresponding feature maps in the encoder network to refine the segmentation labels. Furthermore, the LRN \cite{islam2017label} supervises the predictions at different stages by defining a loss function for each stage.

Lin et al. \cite{lin2016refinenet} devised a multi-path refinement network, called RefineNet, for semantic segmentation. The cascaded architecture exploits multi-scale features from different stages of ResNet \cite{he2016deep} and conveys them into different stages of the RefineNet block via long-range skip connections. The RefineNet block is applied to upsample features maps and to recover the decreased resolution through local residual connections and chained residual pooling. The long-range skip connections are used to integrate information from the coarse high-level deep layers and the fine low-level shallow layers to produce high-resolution semantic feature maps; thus, the gradient can be directly propagated to the inputs, preventing gradient vanishing and explosion.

In conclusion, skip connection structure merges hierarchical cross-layer features to improve the segmentation performance, and the gradient can be propagated backward along both the skip path and the cascaded original path to prevent gradient vanishing and explosion.

\textbf{Atrous Convolution.} Atrous convolution \cite{chen2014semantic, chen2016deeplab}, also called dilated convolution \cite{chen2017rethinking, yu2015multi}, refers to convolution with an atrous rate. The rate corresponds to the stride with which the input signals are sampled. Thus, standard convolution, with a rate of 1, is a special case of atrous convolution.

The fundamental work on the DeepLab system \cite{chen2014semantic, chen2016deeplab} first proposed the definition of atrous convolution and utilizes atrous convolution to simplify the architecture of FCN \cite{long2015fully}. In this work, atrous convolution is constructed via convolutions with upsampled filters. In an atrous convolution operation, the incoming input feature maps are sampled by enlarging the input stride values, resulting in enlarged field of view of filters and feature responses.

Instead of the atrous convolution with a dilated filter in DeepLab \cite{chen2014semantic, chen2016deeplab}, Yu et al. proposed a special tailored atrous convolution in Dilated-Net \cite{yu2015multi} to obtain multi-scale contextual information. Specifically, the atrous convolution in Dilated-Net is built by recomposing the convolution operator itself with dilation factors and is free from the dilated filters in DeepLab. The dilated convolution operator with different dilation factors can adopt the same filter in different ranges to capture multi-scale context. Moreover, the receptive fields are enlarged exponentially without loss of resolution, whereas the parameters in the network grow linearly.

What can be inferred from the aforementioned three works \cite{chen2014semantic, chen2016deeplab,yu2015multi} is that, atrous convolutions can take control of the field of view of the convolution filters and feature responses without additional computation overhead.

\textbf{Coarse-to-Fine Refinement.} Coarse-to-fine refinement exploits cascade or supplementary structures to refine confidence maps from coarse to fine.

Active template regression (ATR) \cite{liang2015deep}, which directly predicts and locates the structural masks for each label, was proposed for human parsing. The structural outputs consist of the mask template coefficients and the shape parameters. ATR builds the end-to-end relations between the input image and the structural outputs by devising two separate CNNs, i.e., a template network and a shape network. The template coefficients are predicted by the template network with max-pooling to capture the contextual correlations among all label masks. Meanwhile, the shape parameters are predicted by the shape network without max-pooling to maintain the sensitivity to the label mask position. The outputs from the two parallel CNNs provide supplementary information. Thus, the normalized mask of each semantic region is expressed as a linear combination of the learned mask templates and is then refined to a more precise mask with the shape parameters.

Li et al. \cite{li2017not} proposed an end-to-end deep layer cascade (LC) framework to improve the accuracy and speed of semantic segmentation. Specifically, LC treats different layers in the deep network as different stages with difficulty-aware learning. The early lower stages are trained to handle easy regions, while the challenging regions are forward propagated to the subsequent higher stages; thus, the prediction process is coarse to fine. Furthermore, dilated convolutions are used on the propagated regions to reduce the computations.

Similar to the LC framework \cite{li2017not}, Zhou et al. \cite{zhou2017fixed} proposed a cascaded fixed-point model for small organ segmentation in a coarse-to-fine manner. The entire input region is fed into a coarse-scaled network to produce the coarse segmentation mask, based on which a small region is generated via a transformation function. Then, the small region serves as the input of the subsequent fine-scaled network to produce a more accurate segmentation result. The fixed-point model is iteratively optimized by means of the strategy in \cite{li2013fixed}.

Wang et al. \cite{wang2017learning} proposed a weakly supervised model, image descriptions in the wild CNN (IDW-CNN), to improve segmentation performance using object interactions and descriptions. The architecture of IDW-CNN is composed of three components, i.e., the feature extraction procedure, segmentation stream (Seg-stream) and object interaction stream (Int-stream). First, ResNet-101 is used to extract features. The Int-stream takes these features as input to predict the object interaction after producing masked features for all categories and outputting an object-presence probability vector for all categories. The Seg-stream first predicts the coarse segmentation masks for each category and further refines the segmentation results by convolving the segmentation masks with the object-presence probability vector obtained from the Int-stream as the filter.

Luo et al. \cite{luo2017deep} proposed a dual image segmentation (DIS) model to boost the segmentation performance using the image-level tags of the IDW dataset rather than using the object interactions and descriptions in IDW-CNN \cite{wang2017learning}. DIS first utilizes ResNet101 to produce the first feature map and the first feature vector for the segmentation prediction net and the tag classification net, respectively. The tag classification net outputs a tag prediction vector for all categories after two-stage refinement of the first feature vector. Meanwhile, in the segmentation prediction net, the second feature map is generated by calculating the sum of the upsampled first feature vector and the first feature map and is then further refined to obtain the initial segmentation map for all categories. The final segmentation prediction is obtained by refining the initial segmentation map with the tag prediction vector.

Essentially, these CNN crafting tricks optimize deep networks from the following perspectives: tailoring convolution or pooling operation in accordance with specific conditions, and modifying connection structure between different level layers. These tricks are universally applicable to all the three levels of semantic image parsing tasks.

\subsubsection{Integration with the Random Field Model}

Some recent studies \cite{liu2015semantic,chen2014semantic,schwing2015fully} integrate random field models, such as Markov random fields (MRFs) \cite{yang2014clothing} and CRFs \cite{yu2015multi}, into deep learning to capture contextual information and long-term dependencies.

The DeepLab system \cite{chen2014semantic} and DeepLabV2 \cite{chen2016deeplab} integrate fully connected CRFs into CNNs to refine the raw DCNN scores and achieve better segmentation results. Fully connected CRF utilizes the Gaussian CRF potentials \cite{koltun2011efficient} to capture long-range dependencies and treats every pixel as a CRF node to receive unary potentials. However, the CNN is separated from the CRF portions, so the DeepLab system and DeepLabV2 are not trained in an end-to-end manner.

Schwing et al. proposed fully connected deep-structured networks (FCDSs) \cite{schwing2015fully} to jointly train the CNN and CRF. On the basis of the VGG16 network \cite{simonyan2014very}, the FCDS incorporates unary potentials into convolutional features and iteratively passes the error of CRF inference backward into the CNN. However, a CNN typically has millions of parameters while a CRF involves thousands of latent variables. Therefore, the simple integration of CNN with CRF is inefficient.

To alleviate this issue, Liu et al. proposed an end-to-end deep parsing network (DPN) \cite{liu2015semantic} that incorporates high-order relations and a mixture of label contexts into an MRF and enables optimal computation of the MRF in a single forward pass rather than using an iterative algorithm. The DPN models unary terms and pairwise terms by the tailored VGG16 network \cite{simonyan2014very} and additional designed layers, respectively.

\subsubsection{Integration with Recurrent Neural Networks}

Because the CNN \cite{liang2015deep, liang2015human} can extract only neighboring context information through small convolutional filters, it obtains only local information, which limits the classification accuracy of each pixel position. Moreover, CRF can learn only the short-term dependencies of sequence data \cite{yu2015multi, yang2014clothing} due to its own inner structure. Therefore, several works \cite{li2016lstm, liang2016semanticA, byeon2015scene, liang2016semanticB, liang2017interpretable, zhang2017progressively} used recurrent neural networks to simulate the graphical model for context modeling. Applications of recurrent neural network architecture range from 1D sequence data, such as speech and language, to 2D image space \cite{byeon2014texture} and semantic segmentation.

Two-dimensional (2D) LSTM architecture \cite{byeon2015scene} was adapted to consider the sophisticated spatial dependencies of labels for the pixel-level segmentation of large natural scene images. Specifically, 2D LSTM simultaneously performs classification, segmentation and context integration with low computational complexity by neglecting additional processing, such as multi-scale. Each local prediction is synchronously affected by its neighboring contexts and their previous spatial dependencies, which helps to efficiently capture local and global contextual information end-to-end.

Similarly to 2D LSTM \cite{byeon2015scene}, the long short-term memorized context fusion (LSTM-CF) model \cite{li2016lstm} was proposed to fuse 2D contextual information from photometric RGB and depth data. LSTM-CF can handle the challenges of severe occlusions and diverse appearances \cite{eigen2015predicting, gupta2014learning, girshick2014rich, farabet2013learning, reza2016reinforcement} for RGB-D indoor scene labeling. The photometric context is captured by stacking several convolutional layers, while the depth context is achieved by devising one LSTM layer that encodes both short-range and long-range spatial dependencies along the vertical direction. Moreover, another LSTM fusion layer is constructed to integrate the 2D contexts from different channels along the vertical direction to achieve true 2D global context through bi-directional propagation of the fused contexts along the horizontal direction. Finally, the 2D global contextual representation is cascaded with the RGB features extracted by convolutional layers.

\begin{figure*}[t]
\centering
\includegraphics[width= 0.9\linewidth]{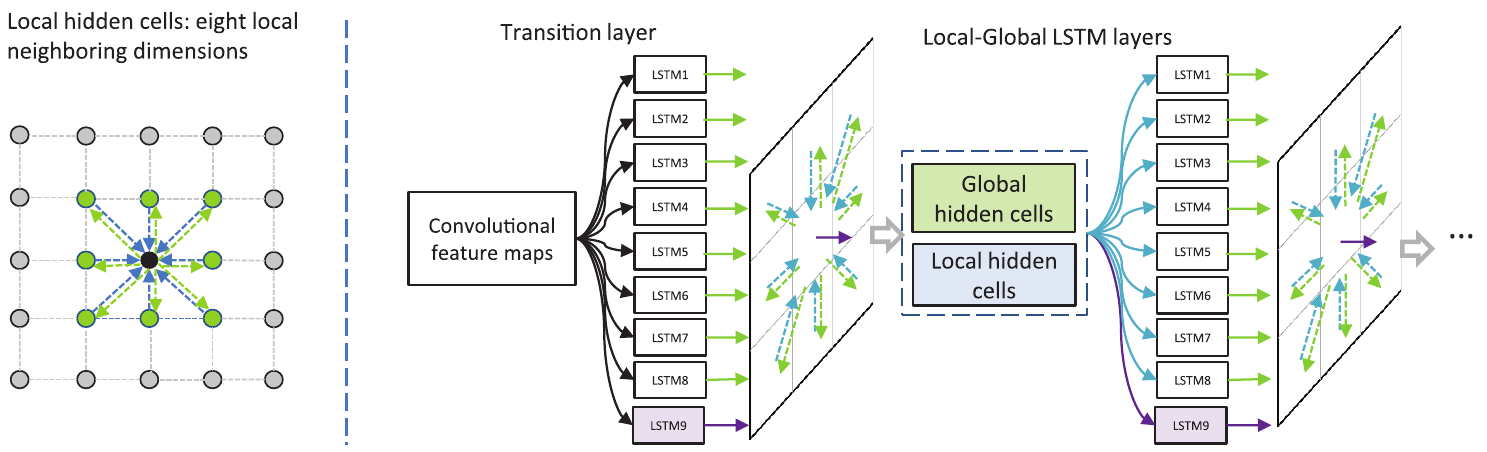}
\caption{Illustration of the LG-LSTM layer. (Figure extracted from \cite{liang2016semanticA})}
\label{fig:9}
\end{figure*}

Local-global LSTM (LG-LSTM) architecture \cite{liang2016semanticA} was developed for end-to-end embedding of local short-distance and global long-distance spatial context into the feature learning over all pixel positions for semantic part segmentation. The local short-distance spatial dependencies of each position in each LG-LSTM layer consist of one depth dimension and eight spatial dimensions (left side of Fig. \ref{fig:9}). The former refers to the hidden cells from the same position in the previous LG-LSTM layer, whereas the spatial dimensions refer to the hidden cells from eight neighborhood positions. Moreover, to capture the global long-distance spatial context (right side of Fig. \ref{fig:9}), in each LG-LSTM layer, the whole hidden cell maps obtained from the previous layer are split into nine grids, each of which covers one part of the image. Then, the global context is obtained by max-pooling operations over each grid. Thus, the features at each position are greatly enhanced by stacking several LG-LSTM layers.

Furthermore, to improve the LG-LSTM architecture \cite{liang2016semanticA}, the graph LSTM \cite{liang2016semanticB} network was built as the generalization of LSTM from sequential data to general graph-structured data. Traditional pixel-wise LSTM structures, e.g., row LSTM \cite{oord2016pixel}, grid LSTM \cite{liang2016semanticA} and diagonal BiLSTM \cite{oord2016pixel, kalchbrenner2015grid}, take fixed-size pixels or patches as physical nodes and capture the context of each node by following a fixed route for different images. By contrast, for each image, graph LSTM constructs a single adaptive graph topology by viewing arbitrary-shaped superpixels as semantically consistent nodes, and the contextual information of each node is obtained along the edges, which represent the spatial relations of the adjacent superpixels.

Another extension of LG-LSTM \cite{liang2016semanticA}, the structure-evolving LSTM model \cite{liang2017interpretable}, was proposed to progressively and stochastically learn interpretable data representations over hierarchal graph structures via LSTM optimization. Structure-evolving LSTM is clearly distinguishable from graph LSTM \cite{liang2016semanticB}, which processes only data with pre-fixed structures. Structure-evolving LSTM stochastically incorporates graph nodes with high compatibilities along the stacked LSTM layers, followed by progressive evolution of the multi-level graph representations from low levels to higher levels, which enables efficient propagation of long-range data dependencies. Moreover, the compatibility of two connected nodes accords with the corresponding LSTM gate outputs in each LSTM layer.

The third extended version of LG-LSTM \cite{liang2016semanticA} is progressively diffused networks (PDNs) \cite{zhang2017progressively}, which unify multi-scale context modeling with deep feature learning for semantic image segmentation. Specifically, PDNs utilize multi-dimensional convolutional LSTMs to construct information diffusion layers, which contribute to diffused information over the learned feature maps. Each LSTM unit is equipped with tailored atrous filters to capture the short-range and long-range context from the neighbors of each site in the feature map.

\subsection{Instance-Level Semantic Segmentation}

Instance-level semantic segmentation has attracted substantial attention \cite{hariharan2014simultaneous, liang2016reversible, liang2015proposal, zhou2017fixed, abtahi2015deep} because increasing practical applications, such as robot task planning \cite{Lin2017Knowledge} and human activity recognition \cite{Lin2016A}, require different objects belonging to the same category to be distinguished. The aforementioned category-level segmentation methods cannot achieve this goal, as illustrated in Fig. \ref{fig:2}. Instance-level semantic segmentation precisely segments each object category and correctly detects all the object instances in one image \cite{he2017mask}, seeking joint object detection and semantic segmentation. Next, we discuss this task from the three aspects: proposal-based framework, multi-task end-to-end module and metric learning embedded model.

\textbf{Proposal-Based Framework.} The first step in the proposal-based framework is to generate proposals, and further processing is required to produce the final segmentations. Most early deep works \cite{hariharan2014simultaneous,hariharan2015hypercolumns,chen2015multi} first adopt a proposal generation method, extract features with tailored CNN architectures, and finally feed the intermediate results into post-processing steps.

Typically, Hariharan et al. \cite{hariharan2014simultaneous} proposed a simultaneous detection and segmentation (SDS) model. This work first generates category-agnostic candidate region proposals via bottom-up multi-scale combinatorial grouping \cite{arbelaez2014multiscale} under the hypothesis that each region proposal, which consists of bounding boxes and initial segmentations, contains one object. On the basis of the proposals, features are extracted from both the bounding boxes and initial foregrounds with two separate tailored R-CNNs \cite{girshick2014rich}. Support vector machines and non-maximum suppression are used to classify region proposals and to refine segmentations, respectively. This work first formulates instance-level semantic segmentation as joint object detection and semantic segmentation. But the computational cost in proposals generation phase is too expensive.

Later research \cite{hariharan2015hypercolumns} follows the same pipeline as that of the SDS model \cite{hariharan2014simultaneous}. The differences among the methods are that 1) the refinement process in the SDS model \cite{hariharan2015hypercolumns} is replaced by hypercolumn-based refinement \cite{hariharan2015hypercolumns} to improve the segmentation accuracy; and 2) in the feature extraction step, this work \cite{hariharan2015hypercolumns} enlarges the bounding boxes set of detections and extracts features from just these bounding boxes without consideration of the region foreground, as in \cite{hariharan2014simultaneous}, which decreases the computational cost.

Recently, Li et al. \cite{li2017instance} presented a novel salient instance segmentation approach that produces salient instance proposals by virtue of salient object contours. Specifically, This work first devised a deep multi-scale refinement network to simultaneously detect salient region and salient object contours. Then, the salient object contours are used to generate salient object proposals, which are futher filtrated by subset optimization algorithm to obtain finer salient instance proposals. The final salient instance segmentation is generated by using CRF model to integrate the saliency mask with instance proposals. This work is a pioneer of joint detection of salient region and salient object contour in a unified framework, and is beneficial for the situation where multiple salient instances are spatially overlapped.

In this proposal-refinement pipeline, proposal generation precedes classification. Apparently, the deep features and large-scale training data play no role in boosting the quality of the generated proposals, therefore, the accuracy of instance segmentation is inherently limited by the quality of the initial object proposals. To resolve the issues, some newly published works \cite{he2017mask, li2016fully} concentrate on unifying the proposal generation and instance segmentation sub-tasks into a single end-to-end framework, and more details are discussed in the below subsection.

\textbf{Multi-Task End-to-End Framework.} Some methods seamlessly integrate the object segmentation of each category and the detection of all object instances into a unified framework, which is beneficial for end-to-end training without supervision in the intermediate stages.

Liang et al. \cite{liang2015proposal} proposed a proposal-free network (PFN) to predict the instance numbers of different categories and each instance segmentation in end-to-end manner. This work \cite{liang2015proposal} directly predicts instance-level masks through bottom-up merging, without requiring object proposals. However, PFN is not suitable for cases with small objects.

Additionally, Liang et al. proposed an alternate novel framework, called reversible recursive instance-level object segmentation (R2-IOS) \cite{liang2016reversible}, which recursively refines object proposals and segmentation masks. R2-IOS contains two significant sub-networks, i.e., the object proposal refinement sub-network and the instance-level object segmentation sub-network, both of which are alternately fed into each other for progressive optimization. The object proposal refinement sub-network reversibly predicts the confidences for all semantic categories and the bounding box offsets to refine the object proposals; meanwhile the instance-level object segmentation sub-network iteratively produces the foreground mask of the dominant object in each proposal. Moreover, one instance-aware denoising auto-encoder is embedded in the instance-level object segmentation sub-network, which helps R2-IOS to distinguish overlapping objects with similar appearance. This work jointly training object proposal refinement and proposal-based segmentation to complement each other, other than works in \cite{hariharan2014simultaneous, hariharan2015hypercolumns, li2017instance}.

Dai et al. \cite{dai2016instanceB} presented multi-task network cascades (MNC), which dissects instance-wise segmentation into three causal sub-tasks respectively accomplished by the three sequential cascaded stages, i.e., distinguishing instances, forecasting masks and categorizing instances. Specifically, MNC first extracts the convolutional features using the stacked convolutional layers. The output is shared among the three following stages. Besides, the outputs from the early stages are also shared among the pursuant stages. This work achieves contemporary state-of-the-art accuracy by transforming complex instance-wise segmentation into three simplified sub-tasks, which, however, has its deficiencies caused by RoIPool \cite{girshick2015fast, he2014spatial}: missing spatial details and repetitive computation among RoIs without sharing.

To alleviate MNC's \cite{dai2016instanceB} issues, FCIS \cite{li2016fully} provides the first fully convolutional end-to-end solution for instance-level semantic segmentation, which highly integrates FCN \cite{long2015fully} for semantic segmentation and InstanceFCN \cite{dai2016instance} for instance mask proposal. Specifically, FCIS is divided into instance mask prediction and classification sub-task. A input image is fed into shallower convolutional layers to produce convolutional representation and further position-sensitive score maps, which are shared between subsequent two sub-networks to exploit the correlation. FCIS is fast, and preserves more spatial details without warping or resizing operations in RoIs. But FCIS has inherent drawbacks at dealing with overlapping instances \cite{he2017mask}.

Recently, a concise general framework, mask R-CNN \cite{he2017mask}, which can simultaneously detect objects in one image and generate a segmentation mask for each instance, as well as being simple to implement and trained, was built for object instance segmentation. In concrete terms, mask R-CNN integrates one mask branch into faster R-CNN \cite{ren2015faster} so that object mask prediction is performed in parallel with the existing branch for bounding box recognition. Mask R-CNN can be generalized to other tasks, such as bounding box object detection and person keypoint detection. More importantly, mask R-CNN transcends all previous state-of-the-art results with its framework's flexibility. However, the accuracy and speed are also restricted by the RPN and RoIPool the same as \cite{girshick2015fast, he2014spatial, ren2015faster}.

\textbf{Metric Learning Embedded Model.} Most recently, the novel research moves towards metric learning embedded deep networks for instance segmentation to measure the likelihood of different elements (e.g., pixels, detections). The distance between different elements is calculated to determine whether these elements belong to the same object instance.

Newell et al.\cite{Newell2016Associative} integrated associative embedding into supervised CNNs for pixel-wise predictions, which view instance segmentation as the joint detection of relevant pixels and their grouping into object instances. Here, the embeddings serve as tags to group detections with similar tags. Specifically, \cite{Newell2016Associative} utilizes a tailored hourglass network to simultaneously produce a detection heatmap and a grouping heatmap for each object category. The detection heatmap affords a detection score at each pixel to predict whether the pixel belongs to the foreground. Meanwhile, the grouping heatmap tags each pixel such that pixels with similar tags are grouped into the same object instance by non-maximum suppression. Besides pixel-wise embeddings \cite{harley2015learning}, this work also engenders pixel-wise detection scores to reduce the output dimension of each pixel.

Coincidentally, Fathi et al. \cite{fathi2017semantic} also manufactured a deep metric learning method to further improve the performance of instance-wise segmentation. Specifically, a fully convolutional scoring model is first adopted to compute the seediness score of each pixel, which estimates the representativeness of the pixel comparing with other pixels in the same instance. Pixels with top seediness score serve as seed points. Then, the distance between the seed points are learned via a deep embedding model, which represents likelihood of two pixels. Thus similar pixels are grouped together into the same instance. Different from \cite{Newell2016Associative} using one-dimensional embedding, this work derives multi-dimensional embedding from each pixels, which makes it more appropriate for slender-shape objects.

Generally, these metric learning embedded models are trained end-to-end with fast speed and promising performance. The grouping procedure is based on pairwise constraints \cite{yang2006distance}, not associated with predefined semantic categories. Therefore, such embedding technology maybe become a new tendency for instance-level segmentation.

\subsection{Beyond Segmentation}

The aforementioned segmentation research focuses on segmenting images with different-level configurations, such as category level and instance level. Each configuration assigns the label of the corresponding level for each pixel. In this section, we discuss beyond segmentation methods, which considers the implicit high-level hierarchical information in the image, such as the geometric information \cite{peng2016geometric}, the relations between objects \cite{wang2017learning}, and the structural information \cite{lin2016deep}, in addition to the aforementioned pixel-wise segmentation. This high-level information improves the image segmentation performance.

\begin{figure}[t]
\centering
\includegraphics[width= \linewidth]{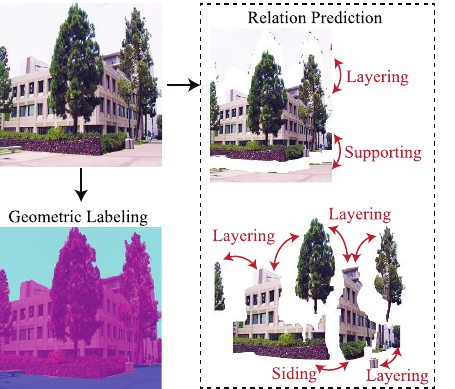}
\caption{Illustration of the geometric scene parsing. (Figure extracted from \cite{peng2016geometric})}
\label{fig:12}
\end{figure}

Peng et al. \cite{peng2016geometric} proposed hierarchical LSTM (H-LSTM) to exploit data from the perspective of geometric attributes and geometric relations, as shown in Fig. \ref{fig:12}. Specifically, H-LSTM simultaneously outputs the segmentation of geometric attributes (e.g., sky, ground) and geometric interaction relations (e.g., layering, supporting) through the pixel LSTM (P-LSTM) sub-network and the multi-scale super-pixel LSTM (MS-LSTM) sub-network, respectively. P-LSTM captures local contextual information to segment geometric attributes; meanwhile, MS-LSTM extracts multi-scale super-pixel representations to categorize geometric interaction relations between adjacent attributes. MS-LSTM shares basic convolutional layers with P-LSTM, which means attribute segmentation and relation categorization benefit from each other.

The major obstacles in beyond segmentation research are the ambiguity of the image hierarchical representations and the rarity of elaborative manually annotated datasets. To alleviate these issues, some \cite{lin2016deep, wang2017learning} introduced top-down information (e.g., hierarchical object structure, object interactions) from image descriptions.

\begin{figure*}[t]
\centering
\includegraphics[width= \linewidth]{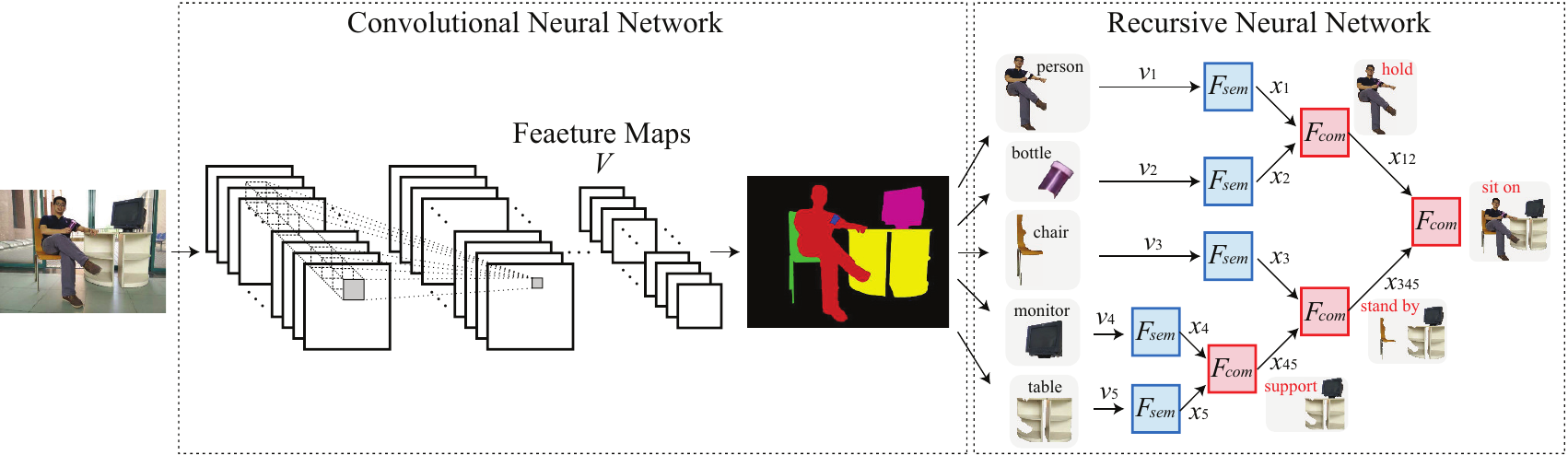}
\caption{The detailed CNN-RNN architecture. (Figure extracted from \cite{lin2016deep})}
\label{fig:11}
\end{figure*}

Lin et al. \cite{lin2016deep} proposed the deep-structured CNN-RNN model by integrating a CNN \cite{long2015fully, krizhevsky2012imagenet} and RNN \cite{socher2010learning}, which can recursively learn the representations in a semantically and structurally coherent way, as shown in Fig. \ref{fig:11}. The CNN layer-wise extracts the feature maps of semantic objects from the input scene image (i.e., semantic segmentation results). Then, the feature maps are fed into the RNN to generate the hierarchically structured configuration (i.e., the hierarchical object structure and the object interaction relations), as shown in Fig. \ref{fig:3}. The CNN-RNN model \cite{lin2016deep} discovers structural scene configurations from the image descriptions \cite{karpathy2015deep,xu2014tell} following the work of \cite{miller1990introduction,socher2013parsing} and is trained in a weakly supervised manner, which avoids the need for elaborate manual annotations. Furthermore, the expectation-maximization method, which alternates between latent label prediction subject to the weak annotation constraints and optimization of the network parameter, is used to train the model.

Inspired by CNN-RNN \cite{lin2016deep}, IDW-CNN \cite{wang2017learning} also exploits the image descriptions \cite{karpathy2015deep, xu2014tell} to capture top-down information, which further improves the image parsing performance. Wang et al. \cite{wang2017learning} designed an elaborate CNN to jointly train IDW and a subsistent image segmentation dataset. IDW dataset are raw: 1) Images are only captioned with raw sentences without pixel-wise annotation; and 2) Images and their descriptions are automatically downloaded from the Internet without subsequent manual post-processing (e.g., cleaning, refinement). IDW contributes useful object interactions to improve segmentation performance; consequently, the precise object segmentation results from the subsistent dataset also benefit object-interaction extractions in IDW. Thus, knowledge from different dataset sources can be fully explored and transferred to improve performance.

\section{Datasets and Evaluation Metrics}

Public datasets and relevant evaluation metrics form the foundation for improving the algorithms. The emergence of big data has driven the development of datasets and relevant evaluation metrics in the field of deep representation for semantic image segmentation, which require large-scale datasets for training. Thus, in this section, we describe these well-known public datasets and evaluation metrics in detail.

\subsection{Datasets}

Table \ref{tab:2} compares the well-known public datasets. According to the source of the image, the semantic segmentation datasets can be divided into RGB(2D) data and RGB-D(3D) data, as indicated by the "2D / 3D" term.

\doublerulesep 0.1pt
\begin{table*}[t]
\renewcommand\arraystretch{1.5}

\begin{footnotesize}
\begin{center}
\resizebox{\textwidth}{23mm}{
\begin{tabular}{c|cccccc}
\hline\hline\noalign{\smallskip}
    Datasets  &\# of images   & \# of training images  &\# of testing images &2D/3D image &indoor/outdoor scene  & \# of categories   \\
\noalign{\smallskip} \hline
   PASCAL VOC 2012 & 9,993 & 4,997  & 4,996 &2D &both  & 20 \\
   PASCAL VOC 2007 & 7,062 & 3,531  & 3,531 &2D &both  & 20 \\
   PASCAL Part  &19,740  &10,103   &9,637   &2D    &both   & 88 \\
   ILSVRC 2016  &25,562    &20,210   &3,352   &2D    &both   &150 \\
   MS COCO      &328,000  &$-$    &$-$  &2D   &both &91 \\

   SIFT Flow   &2,688    &$-$       &$-$   &2D    &outdoor &33 \\
   NYUDv2     &1,449        &$-$      &$-$   &3D   &indoor   &40 \\
   SUN RGB-D   &10,355   &$-$   &$-$   &3D   &indoor   &19 \\
   Fashionista  &685   &456   &229   &2D   &both   &56 \\
   ATR        &7,700   &6,000   &1,000   &2D     &both   &18 \\
   CityScapes   &5,000  &3,475  &1,525  &2D   &outdoor &30 \\

\hline\hline
\end{tabular}}
\caption{Comparison of the semantic segmentation datasets. ``\#'' is short for ``number''. ``$-$'' means the value cannot be found in the literature.} \label{tab:2}
\end{center}
\end{footnotesize}
\end{table*}

\textit{For category-level / instance-level semantic segmentation}, the widely used datasets include PASCAL VOC, PASCAL-Part, ILSVRC 2016, MS COCO, SIFT Flow, NYUDv2, SUN RGB-D, ATR, and Fashionista.

\begin{itemize}
\item \textbf{PASCAL VOC}. The PASCAL VOC dataset \cite{everingham2010pascal} is part of the PASCAL Visual Object Classes (VOC) Challenge organized annually from 2005 to 2012. VOC data have been accepted annually for five main tasks: classification, detection, segmentation, action classification and large-scale recognition. The segmentation task was first introduced in 2007. The dataset is utilized for both category-level and instance-level segmentation. Table \ref{tab:2} lists VOC 2007 and VOC 2012, which are the most frequently used VOC datasets.

\item \textbf{PASCAL-Part Dataset}. This dataset \cite{chen2014detect} contains additional annotations for PASCAL VOC 2010, which provides segmentation masks for each body part of an object.

\item \textbf{ILSVRC 2016}.
The ImageNet Large Scale Visual Recognition Challenge 2016 (ILSVRC 2016) \cite{russakovsky2015imagenet}, organized by the MIT CSAIL Vision Group, is well-known for the image classification task, and it first introduced a scene parsing task in 2016. The dataset for this scene parsing task is the complete ADE20K Dataset \cite{zhou2016semantic}, which contains more than 20K scene-centric images exhaustively annotated with object instances and object parts. Thus, the dataset is used for both semantic instance-level segmentation and category-level segmentation. In particular, the distribution of objects occurring in the images is non-uniform, which simulates daily real-world scenes.

\item \textbf{MS COCO}. The Microsoft Common Objects in COntext (MS COCO) dataset contains 91 common object categories in the version released in 2015 \cite{lin2015microsoft} and 80 categories in the 2014 version \cite{lin2014microsoft}. Distinct from others datasets, MS COCO contains considerably more object instances per image, which may help to exploit contextual information. MS COCO is now a widely used benchmark dataset for category-level and instance-level semantic segmentation.

\item \textbf{SIFT Flow}. The SIFT Flow dataset \cite{liu2009nonparametric} was thoroughly labeled by LabelMe users with 33 semantic categories, 3 geometric categories (i.e., ground, vertical, and sky) and 4 interaction relation labels (i.e., layering, supporting, siding and affinity). The dataset is appropriate for category-level segmentation, and it was later transformed for image geometric parsing.

\item \textbf{NYUDv2}. NYUDv2 \cite{silberman2012indoor} is an RGB-D dataset \cite{gupta2013perceptual} and can be used for both category-level and instance-level segmentation. Additionally, it contains labeled structural support relationships for support relation classification.

\item \textbf{SUN RGB-D}. SUN RGB-D \cite{song2015sun} is the largest RGB-D dataset currently available. The dataset combines most of the previous datasets, such as NYUDv2 \cite{silberman2012indoor}, Berkeley B3DO \cite{janoch2013category}, and SUN3D \cite{xiao2013sun3d}, as well as 3943 newly captured RGB-D images \cite{song2015sun}. Currently, the SUN RGB-D dataset is designed for only category-level semantic segmentation.

\item \textbf{Fashionista}.The Fashionista dataset \cite{yamaguchi2012parsing}, collected from chictopia.com, is designed for clothes parsing and contains 56 different clothing items.Thus, the dataset is tailored for category-level segmentation.

\item \textbf{ATR}. The ATR dataset \cite{liang2015deep}, also used for category-level segmentation, combines four human parsing datasets: Fashionista \cite{yamaguchi2012parsing}, Colorful Fashion Parsing Data (CFPD) \cite{liu2014fashion}, Daily Photos \cite{dong2013deformable} and the Human Parsing in the Wild (HPW) datasets. The labels of the Fashionista and CFPD datasets are merged into 18 categories, and the HPW dataset is newly annotated \cite{liang2015deep}.

\item \textbf{CityScapes}.The CityScapes dataset \cite{cordts2016cityscapes} focuses on both category-level and instance-level segmentation of urban street scenes. It provides 5,000 fine annotations, i.e., individual annotations of single instances, and 20,000 coarse annotations, which cover individual objects with marked polygons.

\end{itemize}

\textit{Semantic image parsing} mainly contains structured semantic parsing \cite{lin2016deep} and geometric parsing \cite{peng2016geometric}. However, to the best of our knowledge, there is no specific subsistent dataset for image parsing. Structured semantic parsing requires not only the segmentation of objects in an image but also hierarchical prediction of semantic objects with object interaction relations. Therefore, the requisite dataset is distinct from the datasets used for semantic segmentation tasks. The work in \cite{lin2016deep} constructed a dataset for this task on the basis of the existing dataset PASCAL VOC 2012. In practice, in addition to utilizing the dataset for category-level semantic segmentation, images, used for constructing the structure and relations, were selected from the PASCAL VOC 2012 segmentation dataset. Furthermore, in contrast to the pixel-wise annotations in this dataset, based on these selected images, \cite{lin2016deep} built image-level annotations by describing each image with several natural language sentences. In addition, each sentence contains objects and the hierarchy with their interaction relations in the image. Similarly, there is no specific dataset to validate the effectiveness of the proposed algorithm for geometric parsing, which simultaneously labels geometric attributes and determines the geometric interaction relations. The work in \cite{peng2016geometric} transformed existing datasets (i.e., SIFT Flow, LM+SUN, and Geometric Context dataset) for use in geometric parsing.

\subsection{Evaluation Metrics}

The performance of pixel-wise segmentation algorithms is commonly evaluated with four metrics \cite{long2015fully}: pixel-wise accuracy, mean accuracy, intersection over union (IoU), and F1 score. Denote $n_{ij}$ as the number of pixels of category $i$ predicted to belong to category $j$, where there are $K$ categories, and let $t_i = \sum_{j} n_{ij} $ be the total number of pixels of category $i$. Then,

\begin{itemize}
\item pixel-wise accuracy: $\sum_i n_{ii} / \sum_i t_i$
\item mean accuracy: $ (1/K) \sum_i n_{ii} / \sum_i t_i $
\item mean IoU: $ (1/K) \sum_i n_{ii} / (t_i + \sum_j n_{ji} - n_{ii}) $
\item F1 score: (Pixel-wise Accuracy + mean IoU) $ / 2$
\end{itemize}

The performance of structured scene parsing algorithms is evaluated with two metrics \cite{lin2016deep}: relation accuracy and structure accuracy. Following \cite{lin2016deep}, the structured scene parsing task is defined as a binary tree, and relation accuracy is computed recursively. Denote the binary tree by T and let $P = {T, T_1, T_2, . . ., T_m}$ be the set of enumerated subtrees (including $T$) of $T$. Each tree consists of objects and relations between different objects, while each leaf only stands for one object. A leaf $T_i$ is considered to be correct if it is of the same object category as that in the ground truth tree. A non-leaf $T_i$ (with two subtrees $T_l$ and $T_r$) is considered to be correct if and only if object categories and relation labels in $T_l$ and $T_r$ are both correctly predicted. Then, the relation accuracy is calculated as $(number\ of\ correct\ subtrees) / (m+1)$. The structure accuracy is a simplification of the relation accuracy that ignores the relation labels when evaluating the correctness of $T$.

\section{Conclusions and Future Work}

In this work, we present a comprehensive review on deep representation learning algorithms for semantic image parsing with a unique perspective. In contrast to other surveys, we review the image parsing models in terms of the development of three-level semantic segmentation from its origins to the most recent, the relatively well-known datasets, and evaluation metrics, including 41 algorithms, 11 datasets and 6 evaluation metrics. We believe that there are several promising research directions for semantic image parsing. The first is multi-task driven semantic parsing, such as \cite{lin2016deep}, which integrates natural language understanding and image parsing. In addition, a large number of training samples are required for deep parsing models, but the collection and annotation of large-scale datasets is elaborative. Therefore, semi-supervised, weakly supervised or unsupervised learning algorithms are another direction to pursue. The third intuitive direction is to transfer image parsing ideas and technologies to the challenging video parsing task.

\noindent

\bibliographystyle{fcsbib2}
\bibliography{semantic}

\end{document}